\documentclass[conference]{IEEEtran}
\IEEEoverridecommandlockouts
\usepackage{cite}
\usepackage{amsmath,amssymb,amsfonts}
\usepackage{algorithmic}
\usepackage{graphicx}
\usepackage{textcomp}
\usepackage{xcolor}
\usepackage{booktabs}
\usepackage{url}
\usepackage{subcaption}

\def\BibTeX{{\rm B\kern-.05em{\sc i\kern-.025em b}\kern-.08em
    T\kern-.1667em\lower.7ex\hbox{E}\kern-.125emX}}
\begin{document}

\title{Multimodal Word Sense Disambiguation in Creative Practice}

\makeatletter
\newcommand{\linebreakand}{%
  \end{@IEEEauthorhalign}
  \hfill\mbox{}\par
  \mbox{}\hfill\begin{@IEEEauthorhalign}
}
\makeatother

\author{
\IEEEauthorblockN{1\textsuperscript{st} Manuel Ladron de Guevara}
\IEEEauthorblockA{\textit{School of Architecture} \\
 \textit{Carnegie Mellon University}\\
Pittsburgh, US \\
manuelr@andrew.cmu.edu} 
\and
\IEEEauthorblockN{2\textsuperscript{nd} Christopher George}
\IEEEauthorblockA{\textit{School of Computer Science} \\
 \textit{Carnegie Mellon University}\\
Pittsburgh, US \\
cmgeorge@andrew.cmu.edu} 
\and
\IEEEauthorblockN{3\textsuperscript{rd} Akshat Gupta}
\IEEEauthorblockA{\textit{School of Electrical Engineering} \\
 \textit{Carnegie Mellon University}\\
Pittsburgh, US \\
akshatgu@andrew.cmu.edu} 
\linebreakand 
\IEEEauthorblockN{4\textsuperscript{th} Daragh Byrne}
\IEEEauthorblockA{\textit{School of Architecture} \\
 \textit{Carnegie Mellon University}\\
Pittsburgh, US \\
daraghb@andrew.cmu.edu} 
\and
\IEEEauthorblockN{5\textsuperscript{th} Ramesh Krishnamurti}
\IEEEauthorblockA{\textit{School of Architecture} \\
 \textit{Carnegie Mellon University}\\
Pittsburgh, US \\
ramesh@andrew.cmu.edu} 
}

\maketitle

\begin{abstract}
Language is ambiguous; many terms and expressions can convey the same idea. This is especially true in creative practice, where ideas and design intents are highly subjective. We present a dataset—Ambiguous Descriptions of Art Images (ADARI)—of contemporary workpieces, which aims to provide a foundational resource for subjective image description and multimodal word disambiguation in the context of creative practice. The dataset contains a total of 240k images labeled with 260k descriptive sentences. It is additionally organized into sub-domains of architecture, art, design, fashion, furniture, product design and technology. In subjective image description, labels do not necessarily correspond to well-defined entities i.e. \textit{cars}, quantitative attributes such as the color \textit{red}, or actions like \textit{playing}. For example, the ambiguous label \textit{dynamic} is a qualitative attribute of an extensive amount of objects and thus, the data's variance is high. To understand this complexity, we analyze the ambiguity and relevance of text with respect to images using the state-of-the-art pre-trained BERT model for sentence classification. We provide a baseline for multi-label classification tasks and demonstrate the potential of multimodal approaches for understanding ambiguity in design intentions. We hope that ADARI dataset and baselines constitute a first step towards subjective label classification.\\
\end{abstract}

\begin{IEEEkeywords}
Multimodal Deep Learning, Word Sense Disambiguation, Subjective Language Understanding, Natural Language, Computer Vision
\end{IEEEkeywords}

\section{Introduction}
One of the main challenges in natural language processing (NLP) is making sense of ambiguous words and sentences. Word sense disambiguation (WSD) is the ability to computationally determine which connotation of a word is implied by its use in a particular context \cite{navigli}. Common problems in WSD are word sense representation, resolution in the distinction of senses, deciding how contexts are defined and the heavy dependency on knowledge external to the text. WSD can be viewed as a classification task, with intended meaning of the words operating as the classes and the words as the objects to be classified \cite{ijcai2018-812}. Thus, supervised learning techniques can be used to solve WSD. However supervised approaches are problematic due to the volume of labelled data required, time-consuming nature of annotation required to provision for the variability in lexical use, that the number of meanings in WSD tasks is in the order of thousands \cite{tripodi-navigli-2019-game}. Yet, most challenging ambiguity-related problems in computational linguistics are easily solved by humans. For instance, in a general context, the sentence \textit{I will read by the lake} has two possible meanings, reading some piece of text somewhere nearby the lake, and reading a text titled \textit{'by the lake'}. State-of-the-art neural machine translation models \cite{tang-etal-2019-encoders, sutskever2014sequence, cho-etal-2014-learning, kalchbrenner-blunsom-2013-recurrent, luong-etal-2015-effective} are capable disambiguating these two alternative meanings.

Creative practice (CP) is a broad term that encompasses those fields that produce creative work. Specifically, we refer as CP to any field that relies on visual mediums and feedback in the creative process. This includes work by designers, architects, visual artists or sculptors for example. In these domains, expressive language used to describe, discuss, disseminate and review work is crucial. Yet, it is especially subjective, inherently ambiguous and challenges clear classification with traditional techniques. For example, imagine a designer says to a colleague that they should \textit{Design the chair with a more dynamic look}; the word \textit{dynamic} conveys the design intent but it can be embodied in intractable number of representations, that is, forms and design variations. To further illustrate this, Fig. \ref{array}\footnote{Creators (row-wise): \textit{SO-IL, Bryanoji Design Studio, Gilles Miller Studio, Igenhoven Architects, Grimshaw Architects, Moarqs, A. Makoto, Gustafson \& Ståhlbom, F. Juhl, C. Kingsnorth, Osko+Deichmann, Dror, P. McDowell, Nike, Campana, Ying Gao, Adidas+A.Taylor, Y. Pardi}. Photography (row-wise): \textit{I. Baan, E. Wonsek and T. Oji, E. Bruzas, Igenhoven Architects, S. Gil, A. Garcia, L. Coleman, Dezeen, Dezeen, Dezeen, Dezeen, Dezeen, N. Nilsen, Dezeen, Dezeen, Dezeen, Dezeen, E. Phillips}.} shows images corresponding to the labels, from left to right: \textit{interesting, simple, organic, different, iconic, minimalist} in the ADARI sub-fields of architecture (top row), furniture (middle row) and fashion (bottom row). This highlights the complex interpretative relationship between descriptions of a workpiece and its form/image. It also leads us the following reserach question: \textit{is it possible to achieve a multimodal WSD through translations of ambiguous text to images?}

This paper introduces an initial dataset, the Ambiguous Descriptions and Art Images (ADARI) dataset of contemporary workpieces, which aims to provide a foundational resource for subjective image description in the context of CP. This has been assembled by collecting articles that include editorial descriptions along with associated images of the creative work. ADARI is an organized dataset of over 240k images and 260k descriptive sentences of visual work provided by the original \textit{creators} or by \textit{curators}. Our findings, explained in more detail in section \ref{amb_rel} and summarized in table \ref{ambiguity_analysis}, show that curators use a more objective language in their descriptions, whereas original creators choose a more ambiguous set of adjectives in their language. 

In addition to the dataset, we provide a baseline for subjective image description and multimodal disambiguation. The baseline architecture is a variant of the CNN-RNN framework proposed by \cite{wang2016cnn} and it is divided into three sections: a convolutional neural network (CNN) image encoder, a subjective label embedding, and a recurrent neural network (RNN) disambiguation decoder. We use a CNN pre-trained ResNet-152 encoder to generate image features, that, along the labels representations, are projected into a joint space. This is the input to the RNN decoder. The multi-label RNN decoder learns a joint low-dimensional image-label embedding to model the semantic relevance between images and labels. The long short-term memory (LSTM) recurrent neurons keep the label context information in their internal memory states, modeling the label co-occurrence dependency in the low-dimensional embedding. The process of generating subjective labels associated to a given image and to a context provides visual meaning to ambiguous words. Fig. \ref{disambiguation} demonstrate the potential of this approach for creative practitioners.

\begin{figure}[ht]
  \centering
  \includegraphics[width=3.49 in]{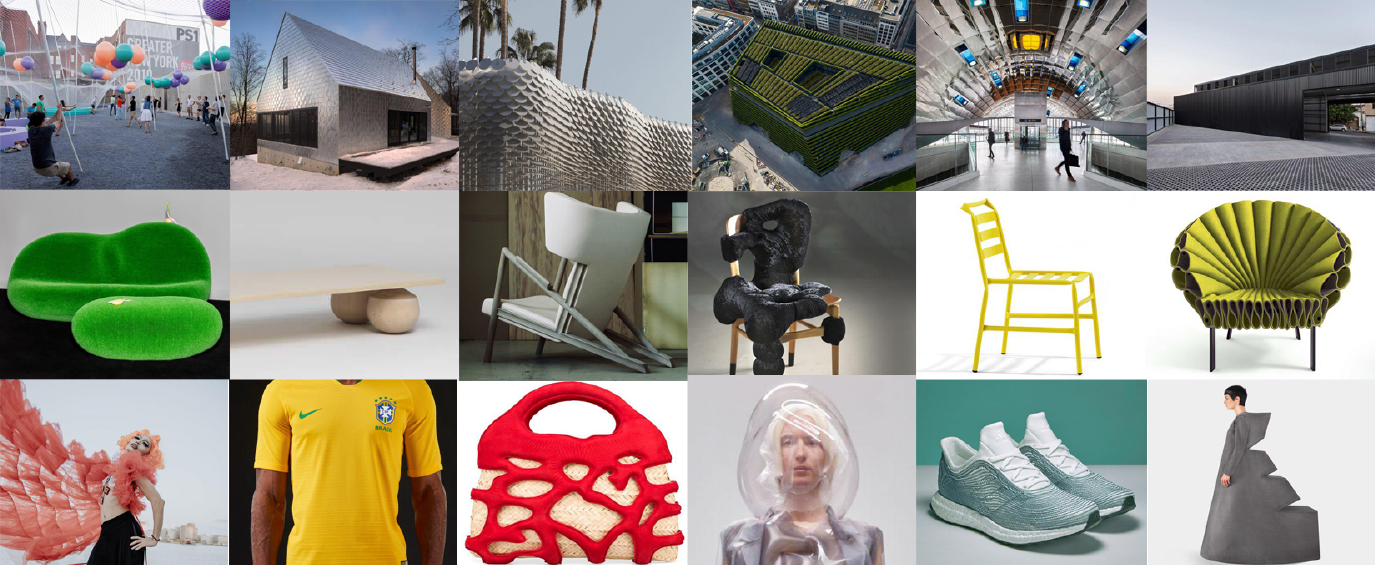}
  \caption{Rows: architecture (top), furniture (middle), fashion (bottom) sub-domains in ADARI. Columns, from left to right: images with label \textit{interesting, simple, organic, different, iconic, minimalist}.}
  \label{array}
\end{figure}

\begin{table*}
  \caption{ADARI raw statistics by sub-field}
  \label{stats}
  \centering
  \scalebox{.90}{
  \begin{tabular}{llllllll}
    \toprule
    \multicolumn{8}{c}{Sub-fields}                   \\
    \cmidrule(r){1-8}
     & Architecture & Art & Design & Fashion & Furniture & Pr. Design & Tech \\
    \midrule
    Samples         & 14227 & 109 & 12805 & 981 & 2564 & 1255 & 1289    \\
    Images          & 124116 & 646 & 83193 & 7131 & 17021 & 7179 & 6774   \\
    Creators sent.   & 60666 & 592 & 26918 & 5341 & 6518 & 6114 & 8825    \\
    Curators sent.  & 75263 & 704 & 40041 & 6345 & 9111 & 7412 & 10178    \\
    Creators adj.    & 123727 & 1001 & 47380 & 8720 & 11812 & 10920 & 15784 \\
    Curators adj.  & 606930 & 3674 & 250297 & 30763 & 48862 & 37810 & 44598 \\
    \bottomrule
  \end{tabular}}
\end{table*}

\section{Related work}
This section reviews existing methods and datasets related to our work, including word sense disambiguiation methods, multimodal learning methods and related multimodal datasets. 

\subsection{Word sense disambiguation}
Navigli describes two main approaches to WSD, supervised and unsupervised WSD, and further distinguishes between knowledge-based or knowledge-rich and corpus-based or knowledge-poor approaches \cite{navigli}. 

Traditional supervised WSD methods such as the work explored by Shen et al. in \cite{shen-etal-2013-coarse} and Iaobacci et al. in \cite{iacobacci-etal-2016-embeddings} concentrate on manual extraction of selected features to train a specific classifier called \textit{word expert} for every target word. Zhong and Ng use Part-Of-Speech (POS) tags and words collocations as contextual features in \cite{zhong-ng-2010-makes} to build support vector machine classifiers for word disambiguation. Among the work that use recent deep neural models, Tang et al. use neural machine translation encoders and decoders, evaluating hidden states and the distributions of attention mechanisms in \cite{tang-etal-2019-encoders}. Reganato et al. explore \textit{all-words} sequence models using bidirectional Long Short-Term Memory (biLSTM) and encoder-decoder architectures to achieve state-of-the-art results in \cite{raganato-etal-2017-neural}.

Tripodi and Pelillo treat the WSD problem as game-theory \cite{tripodi_and_pelillo}, and present a novel semi-supervised technique where each word to be disambiguated is represented as a graph that uses distributional information such as word co-occurrence measures for word similarity and \textit{tf-idf} vectors for sense similarity. This model is reimplemented using dense vectors by Tripodi and Navigli in \cite{tripodi-navigli-2019-game} to achieve significantly improved performance.

Knowledge-based WSD methods exploit the structural properties of a lexical-semantic relationship and usually rely on lexical resources like WordNet \cite{wordnet}. After Navigli and Lapata studied in \cite{navigli_lapata} the potential of enriching such resources with lexical-semantic relations for each word sense, researchers focused on semi- and fully-automatic production of lexical-semantic combinations. Mihalcea and Moldovan created eXtended WordNet in \cite{Mihalcea01extendedwordnet:}, a morphologically and semantically enhanced version of WordNet which aims to provide disambiguated glosses using a semi-automatic approach. Cuadros et al. in \cite{cuadros, cuadros-etal-2012-highlighting} created KnowNet and deep KnowNet, fully-automatic methods for building dense and accurate knowledge bases. More recently, Maru et al. introduced SyntagNet in \cite{maru-etal-2019-syntagnet}, a large-scale manually-curated resource of disambiguated lexical-semantic combinations. Gloss information, which defines a word sense meaning, has been explored by Luo et al. in \cite{luo2018incorporating}, who integrate the context and glosses of the target word into a unified framework in a memory network. Huang et al. construct \textit{context-gloss} pairs fine-tuning pre-trained BERT models in \cite{huang-etal-2019-glossbert}. 

In this paper we present a different approach for WSD, specifically oriented around WSD in creative practice. We differ from the above approaches in a fundamental way: our word senses are not annotated words following semantic relations like WordNet; instead, the sense of a word is represented by learned visual features contained in an image. Visual features are fundamental to many forms of creative practice and communicating design intention, which makes it a compelling domain for exploration. Many prior studies have thoroughly explored design practices to try to understand the subprocesses involved in the design process as a whole \cite{lawson2006designers, mitchell1990logic, cross1989engineering}. For example, Schön and his collaborators introduced new layers of cognitive depth in design modeling, with a special focus on design thinking \cite{schon1984reflective, schon1992kinds}. Schön's conceptual terminology of design as 'interaction with a visual medium' is still relevant to current models of digital design \cite{OXMAN2002135, OXMAN2006229}. Our approach to visually disambiguate the words used in descriptions follows this principle. To the best of our knowledge, this is the first attempt to recourse to visual information to disambiguate the sense of a word within the context of CP.

\begin{figure*}[ht]
  \centering
  \includegraphics[width=7in]{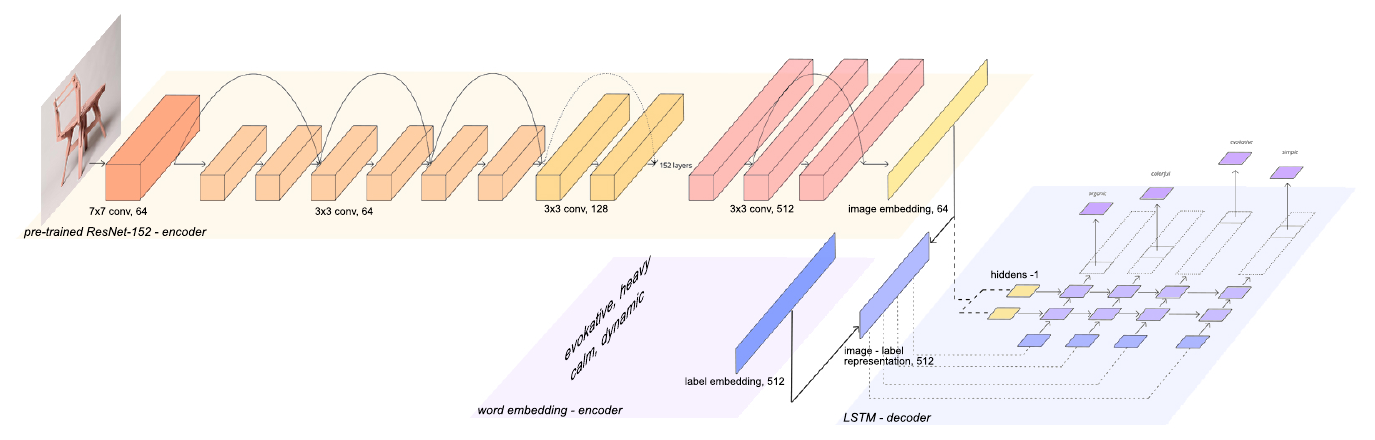}
  \caption{Multimodal WSD encoder-decoder baseline architecture. The network is divided into three section: CNN pretrained ResNet-152 encoder, word embedding encoder, and RNN decoder for multi-label classification}
  \label{baseline}
\end{figure*}

\subsection{Multimodal learning and datasets}
Some applications of multimodal learning frameworks include visual question answering, visual commonsense reasoning or caption-based image retrieval. Lu et al. explore joint representations of image content and natural language in \cite{vilbert}, evaluating their performance in the established vision-and-language tasks mentioned above. They treat visual grounding as a pretrainable and transferable capability using a self-supervised approach. Among supervised approaches, Anderson et al. explore navigation instructions in real environments in \cite{anderson2018vision}. The task of multimodal dialog is explored by Das et al. \cite{das2017visual} while De Vries et al. proposed a modulated visual processing approach for this task in \cite{deVries}. 

Datasets have played a critical role in the development of algorithms and in addressing challenging research problems \cite{lin2014microsoft}. The expansion of object recognition datasets \cite{fei2004learning, griffin2007caltech, dalal2005histograms} had positive effects, such as allowing direct comparison of hundreds of image recognition algorithms as well as generating interest and new approaches to complex problems for the field. Object recognition related datasets can be broadly split into three categories, object classification, object detection and semantic scene labeling. ImageNet, with more than 14 million labeled images \cite{deng2009imagenet}, CIFAR-10 \cite{krizhevsky2009learning} or the popular MNIST \cite{lecun1998mnist} are examples of the first category. The PASCAL VOC datasets \cite{pascal} containing 20 categories with a total of 11000 images, is a representative of the object detection dataset type. The SUN dataset \cite{sun} is probably the most popular for semantic scene labeling. Recently, more complex datasets such as Microsoft COCO \cite{lin2014microsoft} provide the means for a richer individual object segmentation and Conceptual Captions \cite{sharma2018conceptual} which provides a more accurate and richer image captioning. Sanabria et al. released How2, a large-scale multilingual video-text dataset in \cite{sanabria2018how2}, encouraging collaboration between vision, language and speech communities. 

The above work have highly contributed to the growth of the research community, providing new datasets that trigger the investigation of new areas of research. However, even with the increase of interest in applications of machine learning in art, there is no dataset that contains natural descriptions and images of artwork provided by artists or designers.

\section{ADARI dataset}
The Ambiguous Descriptions and Art Images (ADARI) dataset contains 264028 raw sentences and 241982 images in a total of 33230 samples. These were gathered from online articles discussing specific examples of creative visual work. The dataset is divided into seven categories: architecture, art, design, fashion, furniture, product design and technology. These categories establish the \textit{domain} of each image-text pair within the CP \textit{context}. Table \ref{stats} shows the raw counts of samples, images per domain and sentences and adjectives from \textit{creators} and \textit{curators} per domain. Table \ref{ambiguity_analysis} shows that given a visual workpiece, original creators use a more ambiguous language than curators in their descriptions. Unsurprisingly, creators express their subjective ideas in the design process, while curators describe it objectively. Both sources of language are used to create the labels of the dataset.

\paragraph{Data collection}
ADARI is a collection of news articles describing examples of outcomes from creative practice. We harvested the text and images from different online resources for CP content. Each article consists of title, date, text and images. It also belongs to a specific category, for instance, \textit{product design}, establishing the domain. Labeling process is next described in more detail.  

%from \url{http://www.dezeen.com/}, a popular online resource for sharing CP}. Each article consists of title, date, text and images. It also belongs to a specific category, for instance, \textit{product design}, establishing the domain. Labeling process is next described in more detail.  

\paragraph{Data annotation}
ADARI is a self-annotated dataset, as it contains editorial descriptions for each workpiece given by original creators of the work and by curators. The distinction between their descriptions is aided by the magazine's writing style, which contains creators' references between quotes. For each sample, we separately extract the adjectives used by the creator and the curator. To simplify the analysis on how language is used, we generate a bag-of-words (BOW) per domain (furniture, architecture, technology, etc.) and source (creator or curator). To understand the ambiguity of the language, text is analyzed both at the word- and sentence-level.

At a word level we discriminate between relevant and non-relevant and between ambiguous and non-ambiguous adjectives. This process aims to clean non relevant text and to categorize the level of ambiguity. For the purpose of this article, we show the furniture, fashion and wearable technology (sub-domain of technology) ADARI subsets. The classification of ambiguity is a non-trivial exercise. A graphical user interface (GUI) was created to facilitate this labor, and no images were shown in this annotation. The annotators were presented with a list of adjectives, the context, the domain and the annotation criterion. They were then asked to identify whether each adjective is relevant to the context, and if the adjective is deemed ambiguous or not. The criterion is as follows: given the context (design) and domain (furniture), imagine an image of a representative object (chair). If the adjective is able to easily change any attribute of the "external appearance" of the chair, it should be deemed non-ambiguous. We found that descriptive adjectives such as colors, materials or size-related terms tend to be non-ambiguous. Other descriptive adjectives such as \textit{organic} do not express a clear design intent and, therefore, should be classified as ambiguous. Table \ref{ambiguity_analysis} shows the normalized results of the annotation process. Over 500 adjectives per source (creator, curator and both) and per domain (furniture, fashion, wearable technology) have been annotated.

At a sentence level, we discriminate the creator's descriptions between relevant and non-relevant with respect to the images of each article. A GUI facilitates the manual classification of this task. We manually inspected more than 2000 individual samples (over 14000 images and sentences). For each we identified which sentences are related to the images, and therefore, which images are related to the sentences. 

\begin{table*}[t]
  \caption{Normalized ADARI ambiguity and relevance adjective analysis}
  \label{ambiguity_analysis}
  \centering
  \scalebox{.90}{
  \begin{tabular}{cccccccccc }
    \toprule
    \multicolumn{1}{c}{} & \multicolumn{3}{c}{Furniture} & \multicolumn{3}{c}{Fashion} & \multicolumn{3}{c}{Wearable Technology} \\
    \cmidrule(r){2-4}
    \cmidrule(r){5-7}
    \cmidrule(r){8-10}
    Adj source & Creator & Curator & Both & Creator & Curator & Both & Creator & Curator & Both \\
    \midrule
    % \multicolumn{1}{c}{Statistics made by annotators with professional design background}
    % \midrule
    \multicolumn{10}{c}{Annotators with a professional background in design}\\
    \cmidrule(r){1-10}
    Relevance & 0.90 & 0.76 & 0.90 & 0.78 & 0.73 & 0.83 & 0.85 & 0.81 & 0.85   \\
    Ambiguity & 0.79 & 0.63 & 0.75 & 0.80 & 0.59 & 0.67 & 0.70 & 0.58 & 0.61    \\
    \cmidrule(r){1-10}
    \multicolumn{10}{c}{Annotators without a professional background in design}\\
    \cmidrule(r){1-10}
    Relevance & 0.63 & 0.58 & 0.69 & 0.41 & 0.51 & 0.56 & 0.82 & 0.75 & 0.81   \\
    Ambiguity & 0.94 & 0.79 & 0.83 & 0.84 & 0.69 & 0.78 & 0.96 & 0.83 & 0.92    \\
    \bottomrule
  \end{tabular}}
\end{table*}

The ADARI dataset contains images paired with variable length descriptions. Depending on the task, ADARI is structured as follows: 
\begin{itemize}
    \item Word-level: Images are paired a with variable number of quantitative and qualitative adjectives.
    \item Sentence-level: Images are paired a with variable number of sentences subjectively describing the work.
    \item Token-Part of Speech (POS)-Sentence level: Images are paired with tokenized words along POS tags.
\end{itemize}

\section{Experimental settings}

\subsection{Sentence binary classifier}
We use a binary classifier model to distinguish between valid and non-valid sentences. The goal of this experiment is to analyze whether ambiguous creators' descriptions follow specific patterns in the language. The data is sparse and has a high variance as a consequence of the number of creators, in the order of thousands. We use a pretrained state-of-the-art BERT model for sentence classification \cite{Wolf2019HuggingFacesTS}. The model consists of 12-layer BERT with uncased vocabulary and two labels for binary classification.

To find the best performance architecture, we run grid-search over batch sizes of 64, 32, 16 and 8, max sentence length of 64, 32 and 16, and equal-unequal distributions. The latter experiment regularizes the fact that our dataset has 2000 more non-relevant sentences than relevant sentences. The results of these tests can be seen in Fig. \ref{fig:bert_graphs}. As Devlin J. et al. mentioned in \cite{devlin-etal-2019-bert}, the network is more likely to overfit the data as the number of epochs increase. Our best performance on the test set for an equally distributed dataset (Fig. \ref{fig:bert_graphs} a) is 68\%, while achieving 62\% accuracy in the training set. The best accuracy is achieved with a max sentence length of 64. The unequal distribution had similar over-fitting effects, as shown in Fig. \ref{fig:bert_graphs} c. This presents slightly better accuracy, achieving 71\% in the first epoch, and decreasing to 68\% in the third epoch with max sentence length 64. The training set, however, started at 68\% and finished at 82\% accuracy.

\begin{figure*}[t]
  \begin{subfigure}[t]{.4\textwidth}
    \centering
    \includegraphics[width=.8\linewidth]{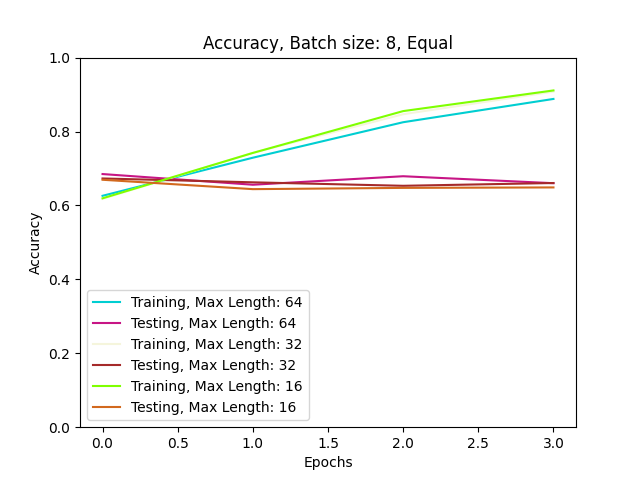}
    \caption{\textbf{Accuracy} of \textit{equally} distributed data with batch size of \textbf{8} and varying max lengths.}
  \end{subfigure}
  \hfill
  \begin{subfigure}[t]{.4\textwidth}
    \centering
    \includegraphics[width=.8\linewidth]{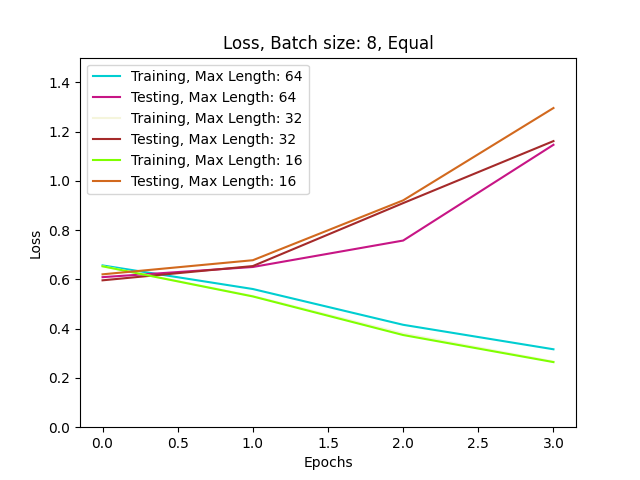}
    \caption{\textbf{Loss} of \textit{equally} distributed data with batch size of \textbf{8} and varying max lengths.}
  \end{subfigure}

  \medskip

  \begin{subfigure}[t]{.4\textwidth}
    \centering
    \includegraphics[width=.8\linewidth]{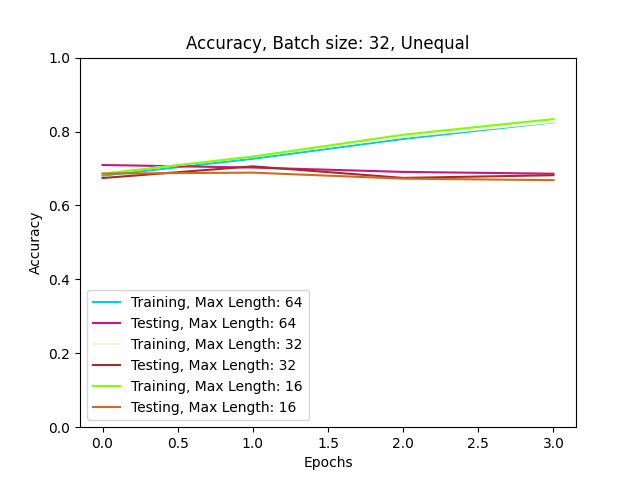}
    \caption{\textbf{Accuracy} of \textit{unequally} distributed data with batch size of \textbf{32} and varying max lengths.}
  \end{subfigure}
  \hfill
  \begin{subfigure}[t]{.4\textwidth}
    \centering
    \includegraphics[width=.8\linewidth]{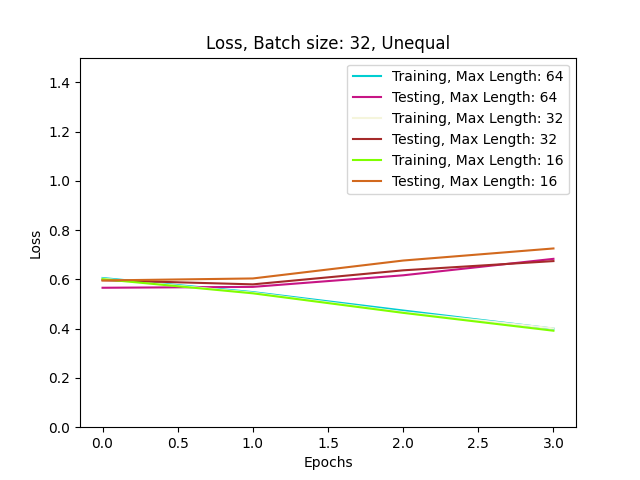}
    \caption{\textbf{Loss} of \textit{unequally} distributed data with batch size of \textbf{32} and varying max lengths.}
  \end{subfigure}
  \caption{BERT Fine-tuning results}
  \label{fig:bert_graphs}
\end{figure*}

\begin{figure*}[t]
  \centering
  \includegraphics[width=\linewidth]{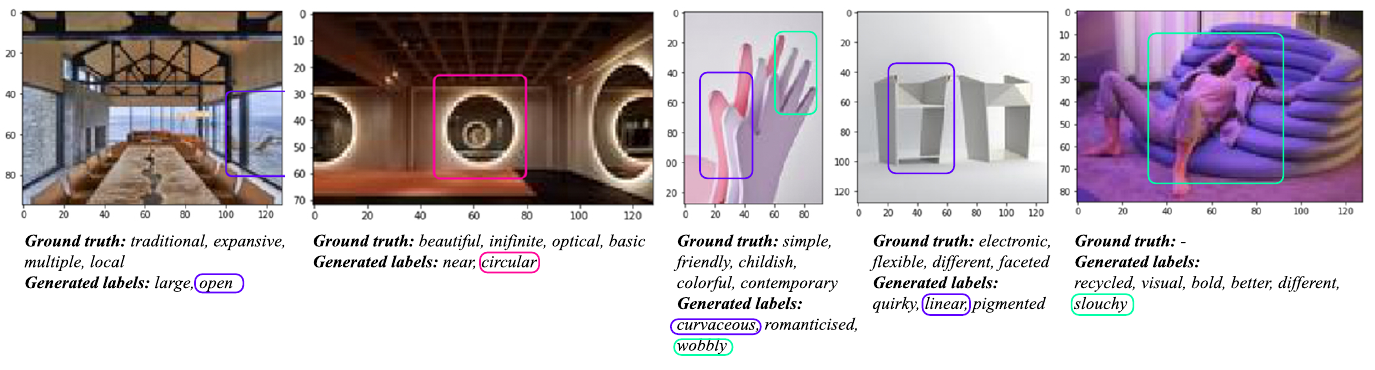}
  \caption{Visual word sense disambiguation, coloured frames show image-words semantic relations}
  \label{disambiguation}
\end{figure*}

\subsection{Multi-label classifier: disambiguation}
\label{amb_rel}
We provide a baseline for subjective image description and multimodal WSD. This framework is composed by 3 parts, as shown in Fig. \ref{baseline}, a CNN image encoder, a label encoder and a recurrent neural network (RNN) disambiguation decoder. The baseline is trained on the ADARI's domain \textit{furniture}, with a random split 80-20 for training and validation sets. The CNN-RNN baseline model performs a multi-label classification via the RNN decoder, which models semantic relations between image and ambiguous descriptions. For the CNN encoder we use ResNet-152 pretrained on ImageNet 2012 classification challenge \cite{deng2009imagenet}, with embedding size of 64 and image input size of 64x64 pixels. Fig. \ref{baseline_results} shows label ranking average precision and coverage loss of these experiments. 

\subsubsection{Baseline architecture}
The proposed CNN-RNN framework for multi-label classification is a variation of the work of Wang et al. in \cite{wang2016cnn}. The CNN encoder extracts semantic representations from images while the RNN decoder models image-labels relationships and label dependency. We work under the assumption that ambiguous descriptions have strong label co-occurrence dependencies as in objective labels \cite{correlative}. The label encoder embeds a sequence of ambiguous words plus a special token \textit{'end'} in a vector of dimension size 64. The decoder consists of an embedding layer of size 64, a 1-layer bidirectional Long Short-Term Memory (biLSTM) of hidden size 512 and a final fully connected layer of vocabulary size. To model image-words dependencies we concatenate the image embedding from the CNN encoder and the ambiguous words sequence embedding, passed as input to the RNN network. The label co-occurrence information is modeled by the RNN hidden states, which take the image representation as input to the hidden states at time step t=0. We performed greedy search to predict labels at each time step. This model achieves the best accuracy of 85.2\% on the test set using 1-biLSTM layer. Fig. \ref{disambiguation} shows the potential of the network to learn semantic relationships between words and images.  

\subsubsection{Training, results and grid-search analysis}
Similar to the binary sentence classification explained, we run grid search to find the best criterion and evaluation metrics, optimizer, architecture and parameters. We experiments are conducted by PyTorch 1.4 and CUDA 10.1 on g4dn.4xlarge Amazon Web Services (AWS) virtual machine, trained during 20 epochs on the ADARI's Furniture domain and evaluated on validation set. 

\paragraph{Criterion and evaluation metrics}
In traditional classification problems, such as binary or multi-class tasks, \textit{accuracy, precision, recall, F-measure and ROC area} are common evaluation metrics that are defined for single label or multi-class classification problems \cite{fawcett2006introduction}. For multi-label classification, however, predictions can be fully or partially correct or fully incorrect. There exists other methods that evaluate ranking such as ranking loss or coverage loss. Ranking Loss evaluates the average proportion of label pairs that are incorrectly ordered for an instance while coverage is the metric that evaluates how far on average a learning algorithm need to go down in the ordered list of prediction to cover all the true labels of an instance. The smaller the coverage, the better the performance \cite{sorower2010literature}. We used coverage error in all experiments. For evaluation, we use label ranking average precision (LRAP) metric. LRAP is the average over each ground truth label assigned to each sample, of the ration of true versus total labels with lower score, where the goal is to give better rank to the labels associated to each sample. Its best value is 1.

\paragraph{Optimizer}
We tested Adam optimizer with an intial learning rate of 1e-3, weight decay of 1e-6 and stochastic gradient descent (SGD) with scheduler of learning rate on plateau in mode \textit{minimum} and patience set to 1. We found that Adam suffer less from fluctuations and achieves a quicker convergence (7-9 epochs). 

\paragraph{Architecture and parameters}
Within the proposed CNN-RNN framework shown in Fig. \ref{baseline}, we explored different LSTM layers, ranging from 1 to 3, and directions, one direction versus two directions. To find the best performance architecture, we set batch size to 64, hidden dimensions of LSTM cells to 512, use Adam optimizer with the same settings as in the above paragraph. We found that 1-biLSTM RNN achieves the best LRPA performance on the test set. We explore RNN hidden dimensions of 512, 256, and 128 using 1-biLSTM architecture and batch size of 64. For batch size performance analysis, we set the RNN hidden dimension to 512, and explore batch sizes of 16, 64 and 128, as shown in Fig. \ref{baseline_results}.

\begin{figure*}[ht]
  \begin{subfigure}[ht]{.35\textwidth}
    \centering
    \includegraphics[width=.8\linewidth]{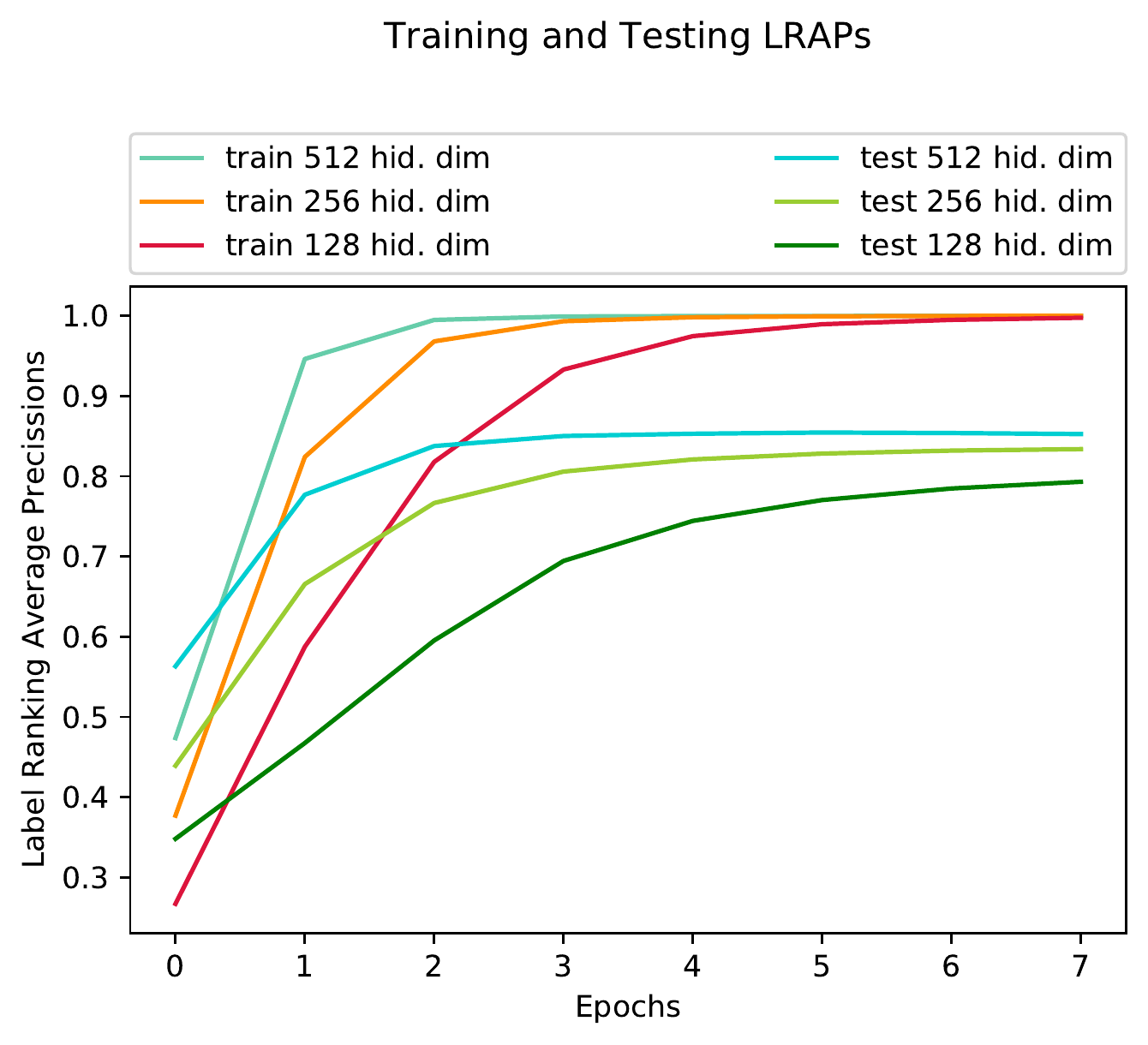}
    \caption{\textbf{Label ranking average precision} with \\batch size of \textbf{64} and varying RNN hidden dimensions.}
  \end{subfigure}
  \hfill
  \begin{subfigure}[ht]{.35\textwidth}
    \centering
    \includegraphics[width=.8\linewidth]{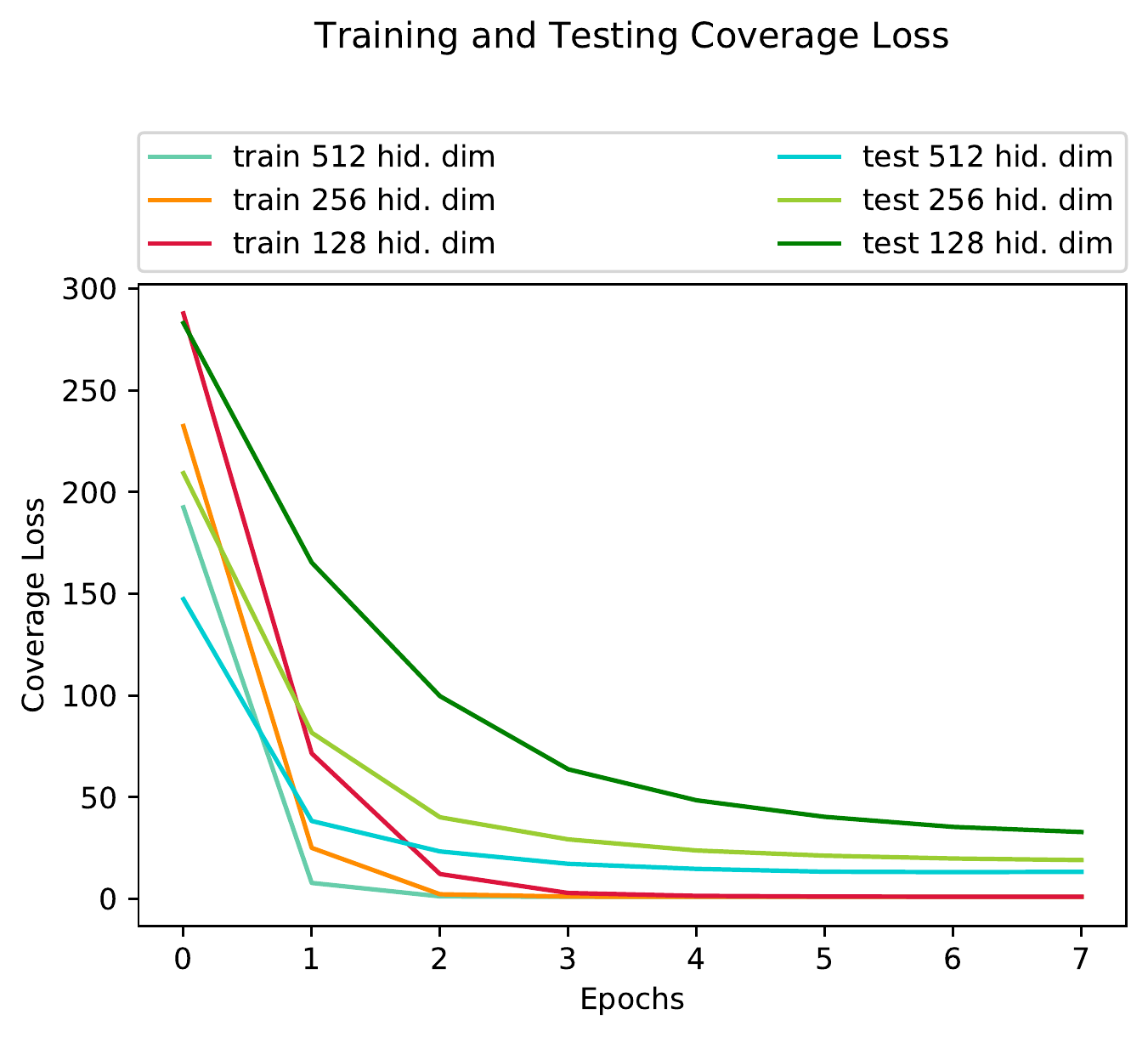}
    \caption{\textbf{Coverage loss} with batch size of \textbf{64} and varying RNN hidden dimensions.}
  \end{subfigure}

  \medskip

  \begin{subfigure}[ht]{.35\textwidth}
    \centering
    \includegraphics[width=.8\linewidth]{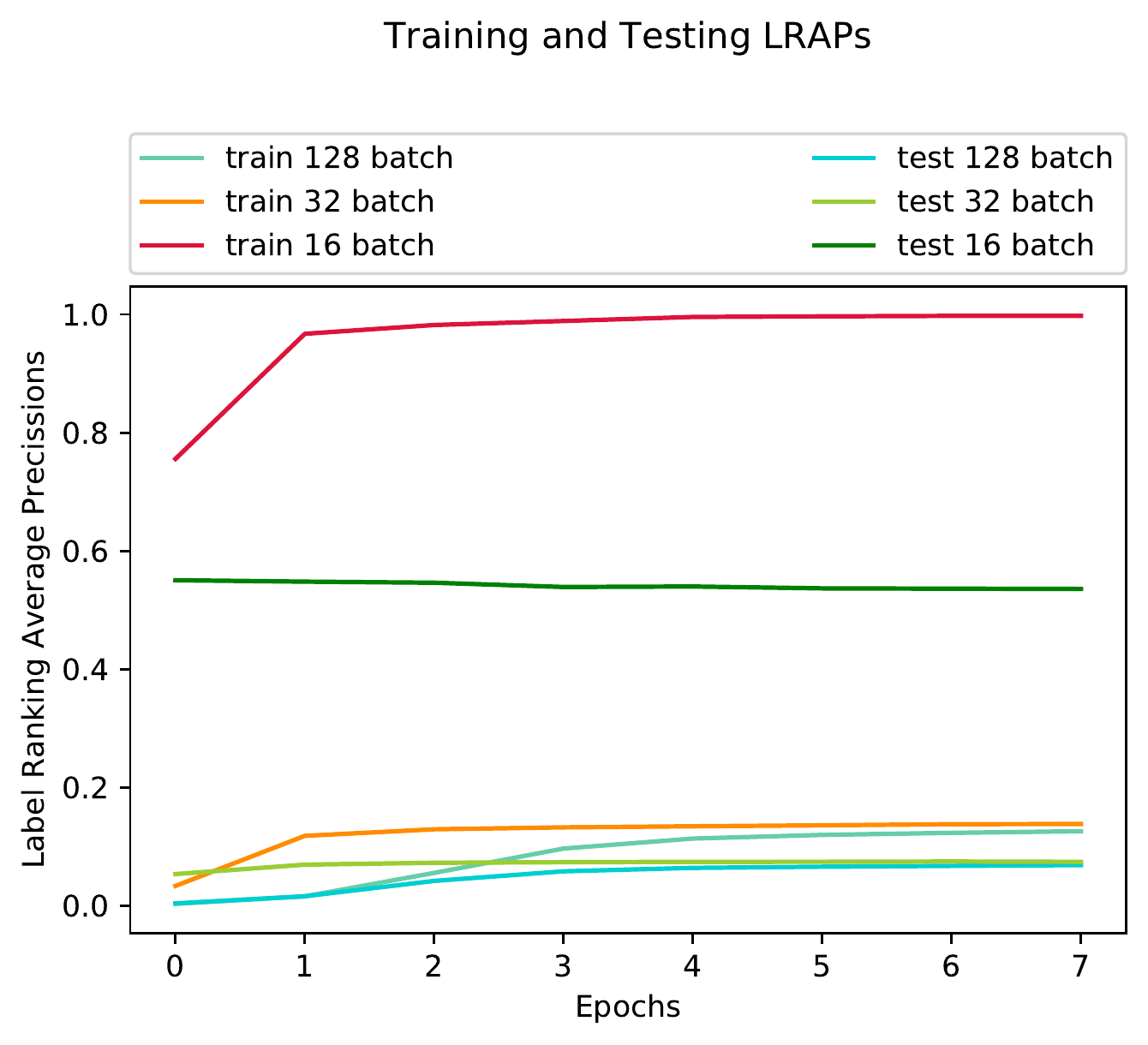}
    \caption{\textbf{Label ranking average precision} with RNN hidden dimension of \textbf{512} and varying batch size.}
  \end{subfigure}
  \hfill
  \begin{subfigure}[ht]{.35\textwidth}
    \centering
    \includegraphics[width=.8\linewidth]{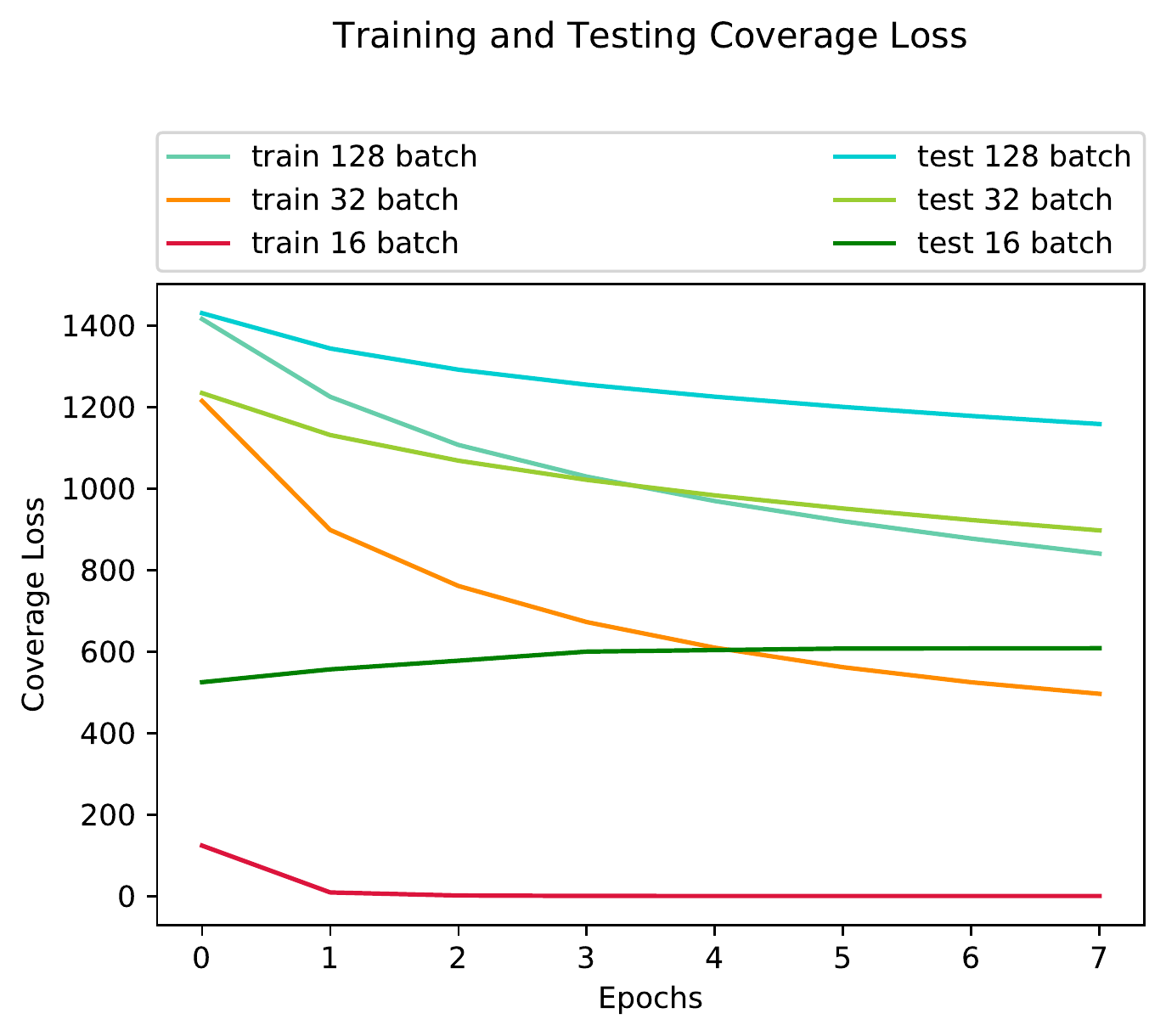}
    \caption{\textbf{Coverage loss} with RNN hidden dimension of \textbf{512} and varying batch size.}
  \end{subfigure}
  \caption{CNN-RNN baseline 1-biLSTM fine-tuning results}
  \label{baseline_results}
\end{figure*}

\section{Conclusion and future work}
We have presented an initial version of the Ambiguous Descriptions and Art Images (ADARI) dataset, which provides a new and foundational resource for subjective image description and visual disambiguation of ambiguous words used by creative practitioners. We have also provided a baseline trained on ADARI's domain \textit{furniture}, which learns the label co-occurrence dependency and helps to visually disambiguate ambiguous words. The benefits of this approach can be seen in two folds: firstly, our CNN-RNN framework can be seen as subjective image description generation. Early results are positive and encouraging for several reasons: the baseline presented is able to detect design nuances in the images that relate to ambiguous words such as “curvaceous”, “wobbly”, “linear”, or “slouchy”, where none of the corresponding images necessarily had those labels applied in the ground truth. This is an indication of a potential approach for understanding ambiguous terms through associations of words-images. Secondly, the potential impact this dataset might have in the research community as an exploratory ground for subjective image descriptioning. However, there are current limitations such as inconsistency for the network to uniformly predict relevant labels. This is partially due to the high variance in images and word senses present in ADARI, and partially due to the baseline architecture's limitations. Future work in relation to the dataset include further expansion, organization and curation of ADARI to support dissemination as a public dataset (it is currently available by request). Future application of such visual disambiguation could help designers in their creation process, as supportive visual communication responsive to natural language.  Further evaluation metrics for subjective descriptioning should be explored. 

\section{Broader impacts}
Research of deep learning frameworks for design processes is underdeveloped with respect to other fields. We have identified a hole that exists for labeled datasets for subjective and ambiguous image descriptions in the context of creative practice.
The motivation for this research arises from the difficult and challenging relationship found in how artists or designers use ambiguous descriptions (language) to communicate about and act upon workpieces (vision). We generated this work with the ambition to further engage and highlight valuable applications for deep learning frameworks within creative fields such as architecture, art or design. We explore ambiguity as means to question conventional design methods and hypothesize that deep learning and natural language processing could revolutionize current design thinking processes (the way we design and make art). This research advocates for a more cohesive position to develop and understand creative practices, avoiding breaking the natural link between human communication and art or design development. %Eventually, understanding the meaning of shape or form through the lens of ambiguity would ground this research and establish the foundations for its applications within the context. 

% \newpage
\bibliographystyle{unsrt}
\bibliography{neurips_2020.bib}

\end{document}